25.6.2026

# How to evaluate clustering with ground truth?

Pasi Fränti
*School of Computing*
*University of Eastern Finland*
Joensuu, Finland

**Abstract:** External indexes can be used for cluster evaluation when ground truth is available. We review the most common external validity indexes focusing on set-matching-based measures. We recommend *centroid index* (CI), because it is an intuitive cluster-level measure with an explainable result. If we need a more fine-tuned, point-level measure, there are more choices. *Pair-set index* (PSI) provides a normalized score which is not biased by cluster sizes. If all points should matter equally, then *clustering accuracy* (ACC) or any other set-matching measure is suitable.

## 1. Introduction

Classification accuracy can be evaluated by comparing the predicted labels against the ground truth labels. In clustering, however, class labels do not exist. The cluster index is an arbitrary number without any correspondence to the ground truth labels. The number of clusters can also differ from that of ground truth, see **Fig. 1**. Evaluating clustering requires, therefore, slightly more attention than evaluating classification.

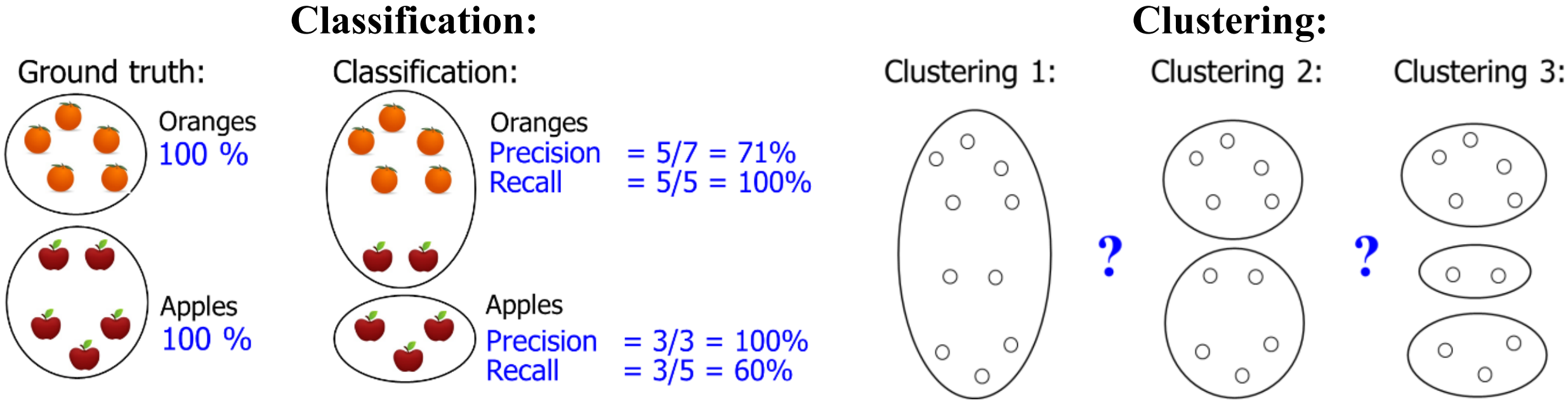

**Figure 1**: Evaluation of clustering differs from that of classification.

According to [1], there are three approaches to evaluate clustering:

1. Application-level performance
2. Internal validation
3. External validation

The first choice is to measure the effect of the clustering in a larger application. However, clustering is just one component, and rarely the most critical. In classical voice biometrics, clustering was used to model the distribution of feature vectors [2], as shown in **Fig. 2**. Random subsampling was found insufficient, and some clustering algorithm was required. However, the exact choice of the algorithm was not critical [3]. It was enough to avoid the poorest algorithm and not use the smallest model size; see **Fig. 3**. For this reason, application-level performance is not very useful for providing deeper insight into the clustering algorithm.

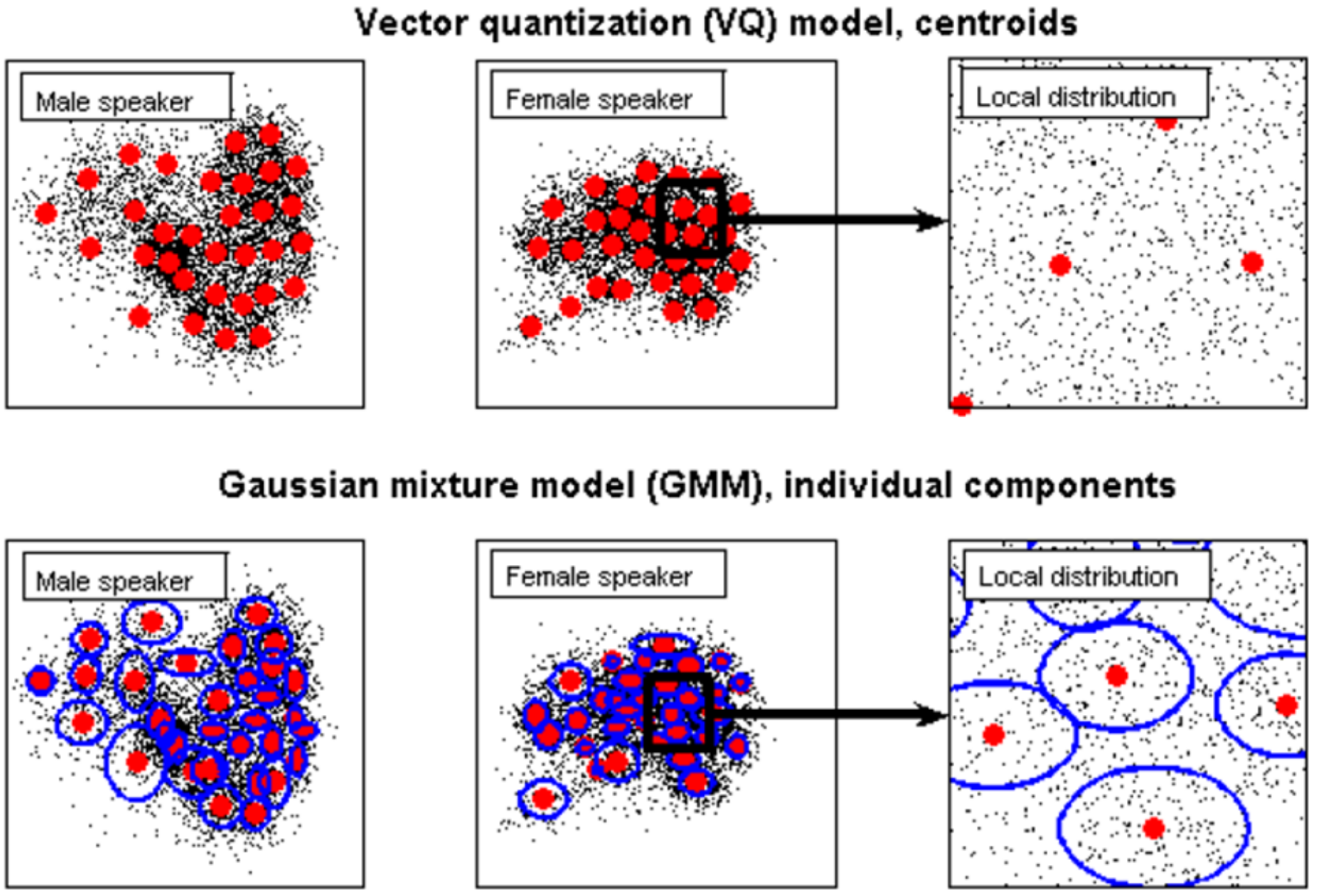

**Figure 2**: The role of clustering here is to model data distribution. Two different models are demonstrated: vector quantization (centroids) and the Gaussian mixture model.

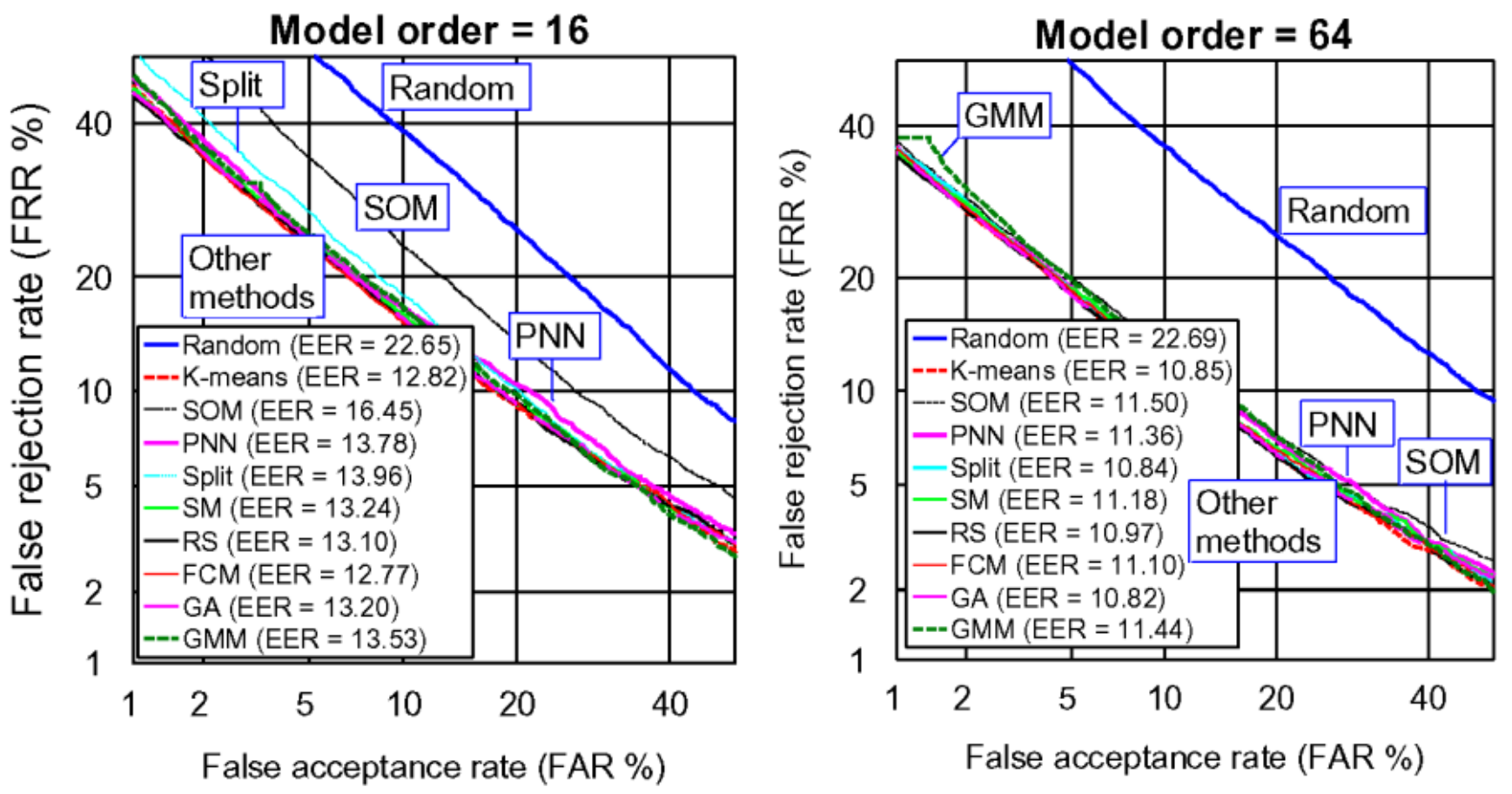

**Figure 3**: The choice of clustering method algorithm has only a marginal effect on the recognition performance. [3]

The second approach is internal validation. It is the easiest as it does not require ground truth or a large application for testing. All we need is a cost function (internal validity index) to optimize. The most common cost function is the sum-of-squared errors (**Fig. 4**):

$$SSE = \sum_{i=1}^{N} dist(x_i, c_{a(i)})^2$$

It measures the squared distance between the data points X={$x_1, x_2, \ldots, x_N$} and their cluster centroids C = {$c_1, c_2, \ldots, c_k$}. A = {$a_1, a_2, \ldots, a_N$} is the cluster assignments of the datapoints so that $a_i \in [1, k]$. *Dist* is the distance function (usually Euclidean). SSE minimizes intra-cluster variance (compactness) directly, and maximizes between-cluster variance indirectly [4].

Finding a good algorithm minimizing SSE is essentially a solved problem. In theory, minimizing SSE is an NP-hard problem, meaning that finding a guaranteed global minimum may take exponential time. In practice, however, we do not need to find the exact global minimum. Typical solution space can have multiple plateaus with virtually equal SSE-value [5]. A good algorithm, such as random swap [6], genetic algorithm [7], or global k-means [8], is likely to find such, and even repeated k-means have a good chance when the clusters overlap [9]. Random swap can find the correct global allocation in $O(Nk^2)$ expected time [6] if the data contains detectable clusters.

Internal validation can also include the number of clusters ($k$) as an unknown variable, which must be estimated as well. Some indices work well in practice, including Silhouette [10], Calinski-Harabasz [11], WB-index [12], and kCE-index [13]. The Davies-Bouldin index [14] is also popular but exhibits unstable behavior and is often inaccurate; therefore, it is not recommended. For an implementation of these and several others, we refer to [15].

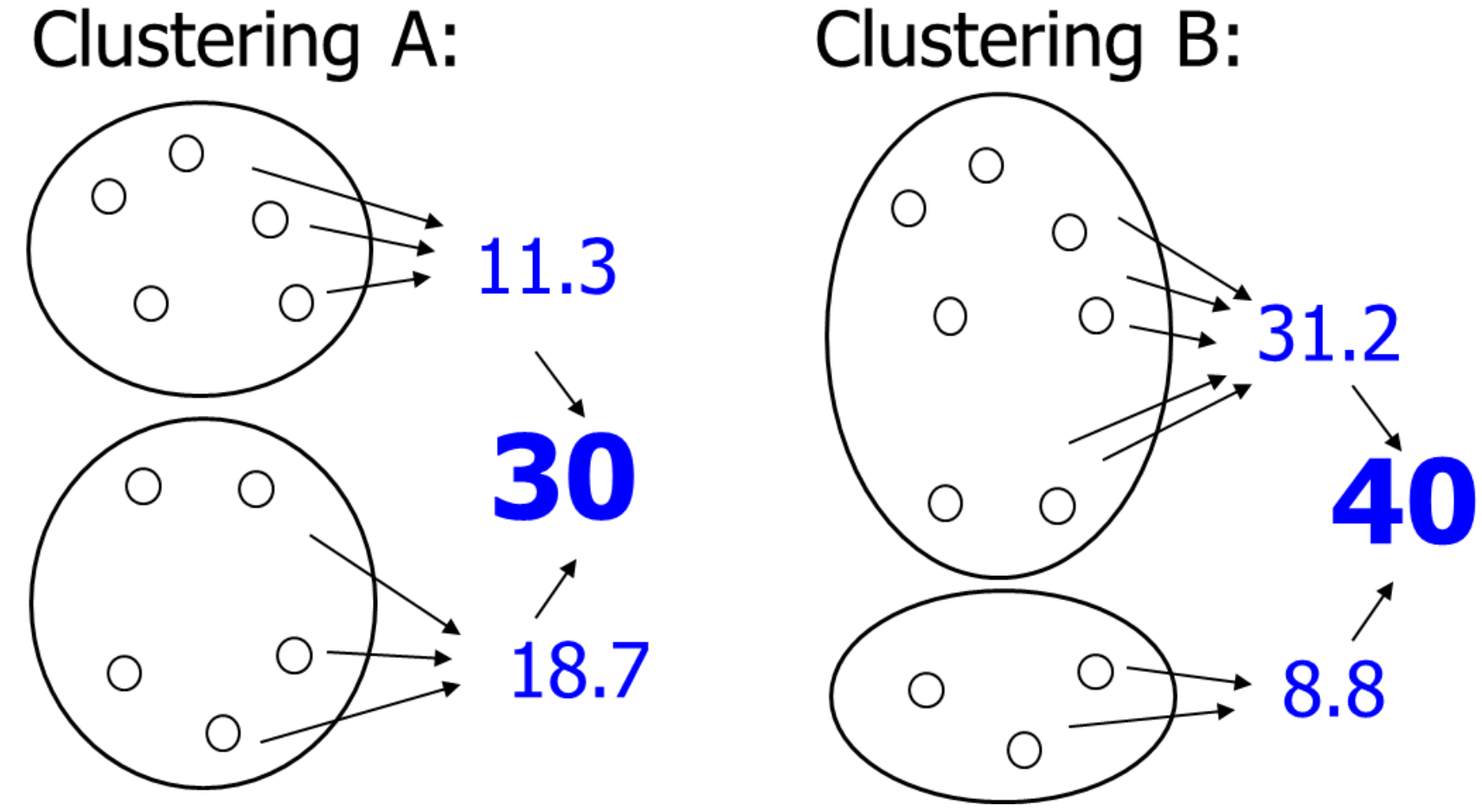

**Figure 4**: Most internal indexes aim at minimizing variance within the clusters. Here, clustering A leads to a smaller SSE than clustering B.

The main advantage of internal indexes is that they can be used directly as the objective function in the optimization algorithm [16, 17]. Their main drawback is that they do not provide a clear, explainable interpretation. We can optimize this value, but cannot know how far the result is from the optimum, or whether the correct clustering structure was found (if it even exists).

To address this issue, artificial datasets with known ground truth have been widely used to evaluate clustering algorithms. The ground truth can be given either as partition labels or cluster centroids. For example, most benchmark datasets in [18] were created by selecting $k$ seed points (cluster centroids) and generating random points around them with a Gaussian distribution. The clustering result can then be compared against the ground truth using an external index.

The most common external indices in the literature are *the adjusted rand index* (ARI) [19], *normalized mutual information* (NMI) [20], and *clustering accuracy* (ACC) [1]. We next review the most common clustering validity indexes. They can be divided roughly into two categories:

1. Pairwise measures
2. Set-matching-based measures

We first briefly cover the first class and then focus on the set-matching-based measures. Note that all the measures can be used to evaluate the similarity of any two clusterings, not only against ground truth.

## 2. Pairwise Measures

Classical *Rand index* [21] measures the consistency of all pairwise datapoints in terms of how they are clustered. If a pair of points is in the same cluster in the ground truth, they are expected to be in the same cluster also in the clustering result. And vice versa: if two points are in different ground truth clusters, they are expected to be clustered into different clusters, see **Fig. 5**.

However, there are a quadratic number of pairs, and most pairs appear in different clusters in the ground truth. The measure is therefore overwhelmed by the agreements of such pairs (*d*) [22]. Consequently, Rand index provides high values (close to 1.0) for almost any reasonable clustering.

Adjusted Rand index [19] normalizes the measure according to the expected value of random clustering. This makes the measure more useful for practical applications, but it still tends to yield relatively high values for clustering results that are far from perfect. Also, if cluster sizes are unbalanced, ARI will primarily reflect the agreement of the biggest clusters [23].

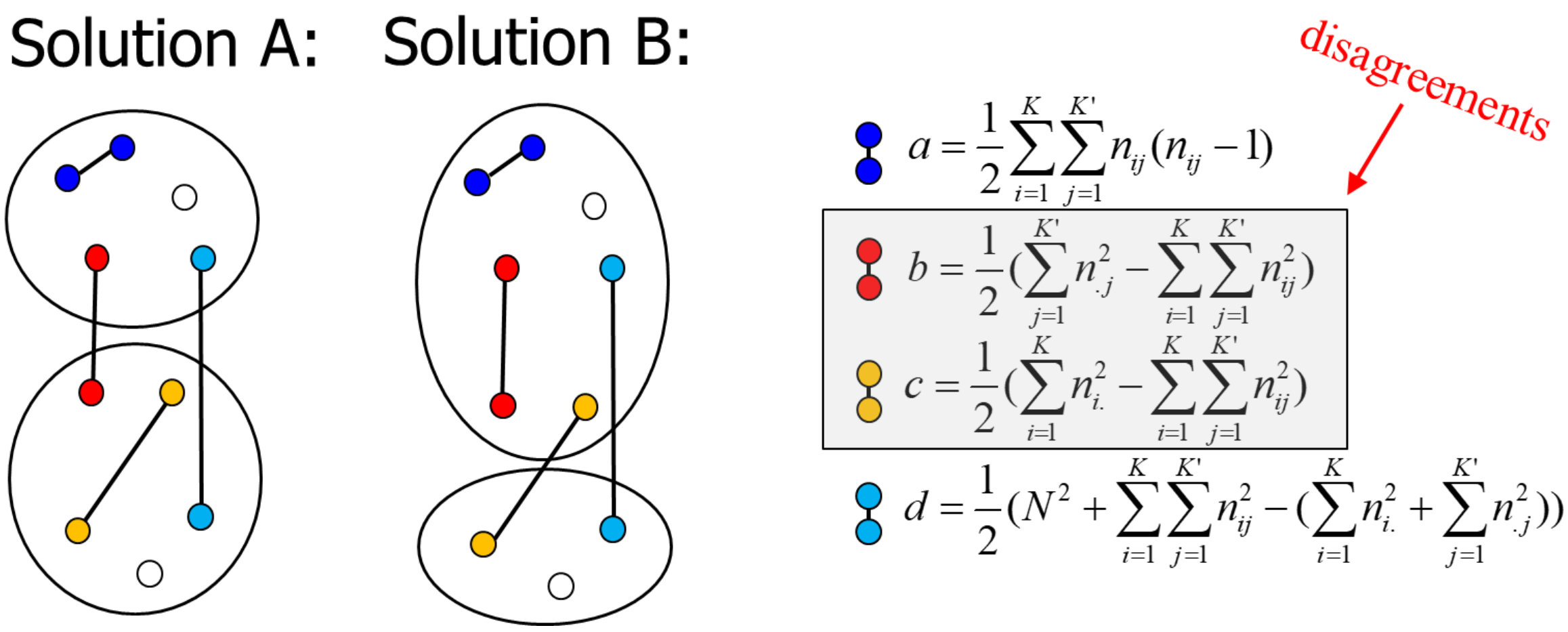

**Figure 5**: Pairwise measure evaluates the consistency of whether a pair of points is clustered in the same way (same cluster or different cluster), both in the ground truth and in the clustering result. Rand index counts the percentage of consistent points: RI = $(a+b)$ / $(a+b+c+d)$. Here $K$ and $K$' are the number of clusters, $n$ and $n$' are the number of points in two clustering solutions A and B.

Information-theoretic measures extend the consistency of a pair of points to a pair of clusters. *Mutual information* (MI) measures how much a pair of clusters shares, quantified by their entropy. The more similar the clusters, the lower the information value. *Normalized mutual information* (NMI) is the most common index in this class, see **Fig. 6** and **Fig. 7**. It was shown to be equivalent to several other information-theoretic indexes, including *adjusted variation of information* ($AVI_S$), *normalized variation of information* ($NVI_S$), and *adjusted mutual information* (AMI) in [24].

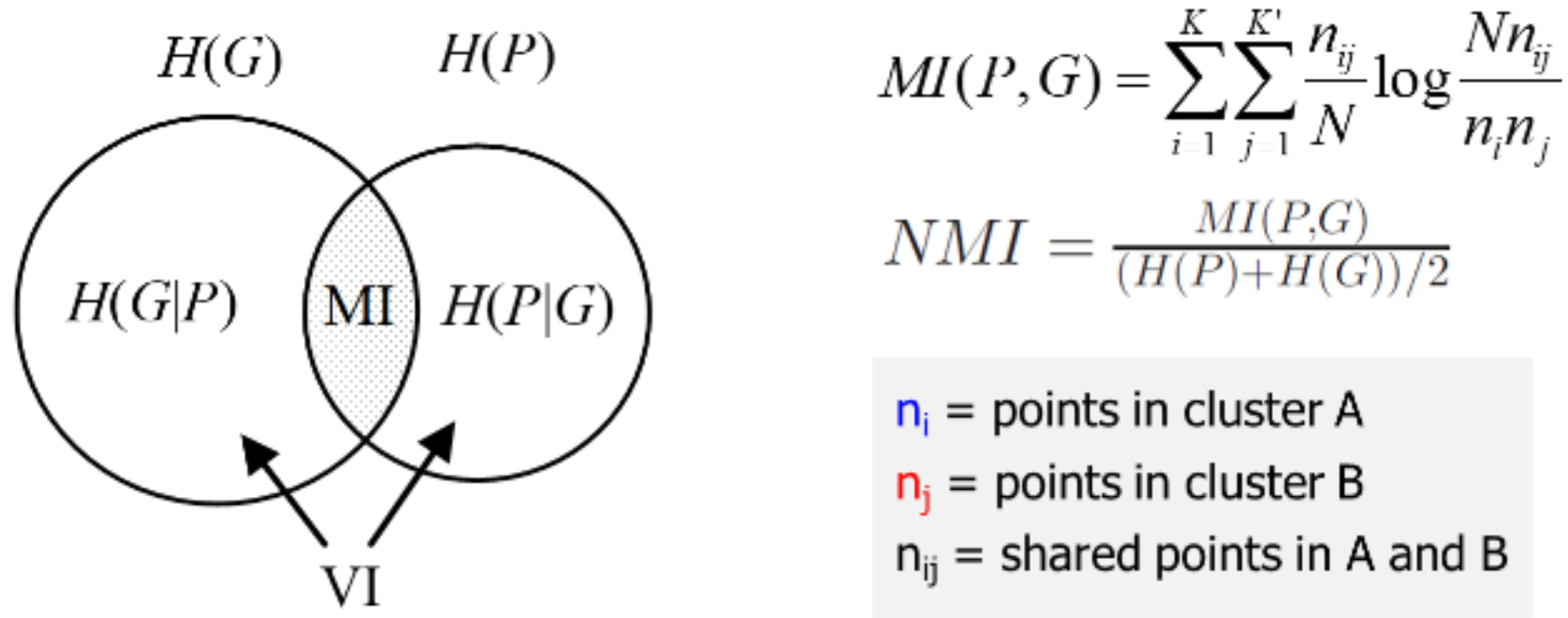

**Figure 6**: NMI measures cluster-level consistency as the shared information value of a pair of clusters relative to the self-entropy of the two clusters. Here H(G|P) and H(P|G) refer to conditional entropy.

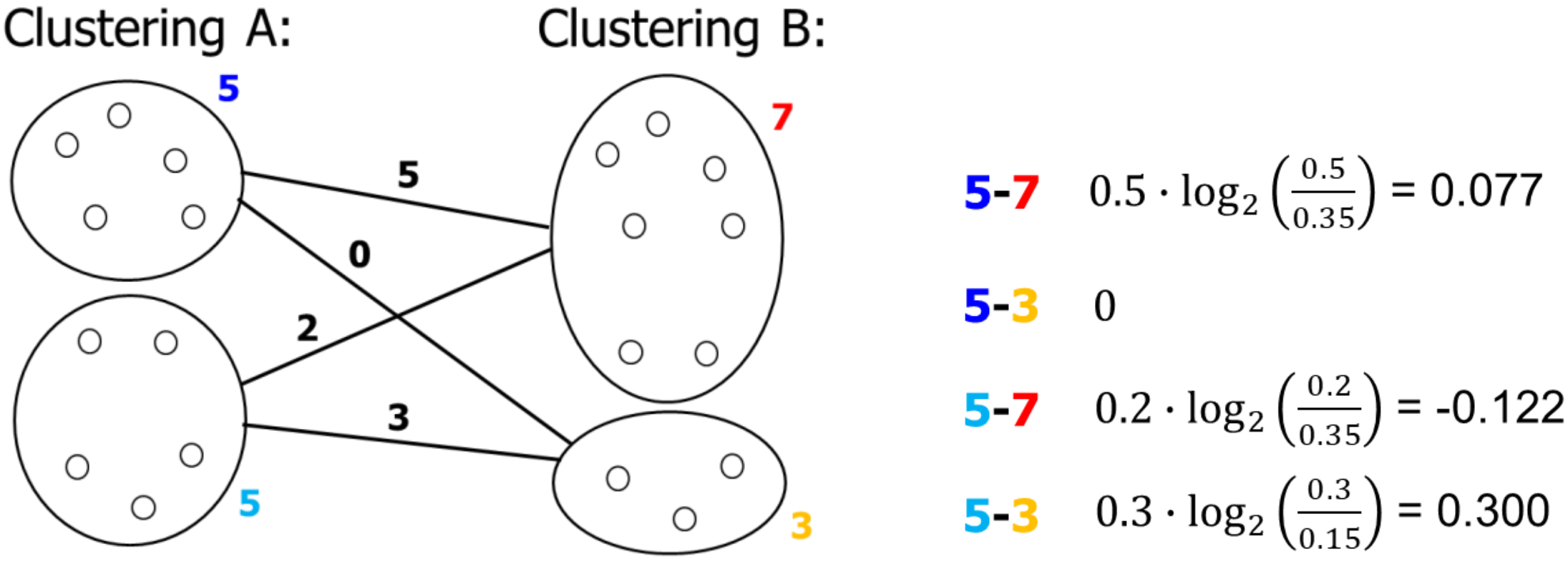

**Figure 7**: Example of measuring NMI between two clusterings, A and B. The number on an edge refers to the number of shared points in the two clusters the edge connects.

## 3. Set-matching-based measures

Instead of comparing every pair of clusters, set-matching measures find the best-matching cluster for each cluster and calculate their similarity [24]. This principle is meaningful as a *cluster* is the main component in a clustering result. There are three design components:

- How to match the clusters?
- How to measure their similarity?
- What about normalization?

Three measures were invented around the same time in early 2000: *normalized van Dongen* (NVD) [25], *Criterion H* (CH) [26], and *clustering accuracy* (ACC). They follow the same principle with only minor differences in their design. ACC is the most popular due to its fancy name and the publicly available MATLAB code by [27], but it was not properly documented until recently in [18]. Several other measures have also been introduced as slight variations of these three. Table 1 summarizes the design choices for some variants, which we discuss in the following subsections.

**Table 1.** Variants of the set-matching-based measures. Here, $A$ refers to the clustering result and $G$ to the ground truth, but they apply to comparing any two clusterings, $A$ and $B$.

| Measure | Year | Ref. | Similarity in… Mapping | Score | Mapping strategy | Normalization |
|---|---|---|---|---|---|---|
| ACC | 2000 | [1] | $\|A \cap G\|$ | $\|A \cap G\|$ | Optimal pairing | $N$ |
| NVD | 2000 | [25] | $\|A \cap G\|$ | $\|A \cap G\|$ | Matching (2-ways) | $2N$ |
| CH | 2001 | [26] | $\|A \cap G\|$ | $\|A \cap G\|$ | Greedy pairing | $N$ |
| Purity | 2011 | [28] | $\|A \cap G\|$ | $\|A \cap G\|$ | Matching ($G \rightarrow A$) | $N$ |
| FM | 2012 | [29] | SD | $\|A\| \times$ SD | Matching ($G \rightarrow A$) | $N$ |
| CI | 2014 | [5] | Nearest | 0/1 | Matching (2-ways) | - |
| CSI | 2014 | [5] | Nearest | Jaccard | Matching (2-ways) | $2N$ |
| CR | 2014 | [30] | Nearest | 0/1 | Greedy pairing | $k$ |
| GCI | 2016 | [31] | $\|A \cap G\|$ | 0/1 | Matching (2-ways) | - |
| PSI | 2016 | [24] | BB | BB | Optimal pairing | $k$ |

### 3.1 Matching

Matching the clusters is the key component. There are two ways to do it:

- Mapping
- Pairing

One-directional mapping takes a cluster from one solution and finds its best match in the other solution. *Purity* and *F measure* (FM) assume that the first solution is ground truth ($G$) and the mapping is from $G$ to $A$. The one-directional mapping may fail if the number of clusters differs. For example, mapping from B to A in **Fig. 8** yields a perfect match when no normalization is applied. For this reason, *normalized van Dongen* (NVD), *centroid index* (CI), *centroid similarity index* (CSI), and *generalized centroid index* (GCI) perform bidirectional mapping and calculate the average (or maximum) of the two mappings as the final score.

The second approach considers matching as a pairing problem. This prevents multiple clusters from being mapped to the same cluster. Some measures require that the number of clusters be equal in both solutions. This is the case with the original ACC measure [32] and the corresponding MATLAB code [27], but the revised version addressed this by creating dummy clusters before pairing [1], as shown in **Fig. 8** (i.e., the two empty clusters in Solution A on the right).

Solving the mapping problem is easy: it is essentially a nearest-neighbor search. Solving the pairing problem is more challenging. CH and CR use greedy pairing, finding the best match for each cluster one by one. ACC and PSI solve the optimal pairing using the Hungarian algorithm. The original algorithm is quite complicated and takes $O(k^3)$ time [33, 34], which is usually not a problem since $k \ll N$. ACC uses the Jonker-Volgenant variant [35] with the same worst-case time complexity, but is much faster for sparse matrices, close to $O(k^2)$ time. For the full story of ACC, we refer to [1], and the software is on GitHub[1] including both the mapping and pairing variants.

[1] https://github.com/uef-machine-learning/ClusterAccuracy

**Figure 8**. Example of mapping (left) and pairing (right). The total number of matched points is (5+10)/20 = 75% in mapping, and 5/10 = 50% in pairing. Mapping tends to find more matches than pairing.

### 3.2 Similarity

A similarity measure is needed to count how much the matched clusters have in common. It is needed for two purposes. First, we need it to select the best match for every cluster. Second, we need it to compute the final score for the measure.

The first (and most common) approach is to count the number of shared points $|A \cap G|$. ACC, NVD, CH, Purity, and GCI use this raw score. However, when cluster sizes are unbalanced, the score is dominated by the largest clusters, as with ARI [23]. This is fine if we want all points to contribute equally. However, if we want all *clusters* to contribute equally, then a normalization step is needed.

The second approach is to apply a set similarity measure to make the score invariant to cluster sizes. CSI uses Jaccard, which divides the intersection size by their union size. PSI uses Braun-Banquet, which divides by the maximum cluster size. FM uses Sorensen Dice, which divides by the average of the cluster sizes; although the normalization effect is canceled later by multiplying the result by the set size |A|. These similarity measures are listed below:

| | | |
|---|---|---|
| Shared points: | S | $= \|A \cap G\|$ |
| Jaccard: | J | $= \|A \cap G\| / \|A \cup G\|$ |
| Braun-Banquet: | BB | $= \|A \cap G\| / \max(\|A\|,\|G\|)$ |
| Sorensen-Dice: | SD | $= \|A \cap G\| \cdot 2 / \|A\|+\|G\|$ |

The third approach is to match the centroids. CI, CSI, and CR find the nearest centroids in the other solution. CI counts the number of unmatched centroids (orphans). CR counts the number of unstable centroids. CSI uses the nearest centroid for matching but applies Jaccard for scoring. GCI is a generalized version of CI; it uses $|A \cap G|$ for the matching, but the number of non-matched centroids as the score.

### 3.3 Normalization

Most measures normalize simply by dividing by the number of points ($N$). In this way, all points contribute equally, and the score is in the range [0,1]. Using set-matching measures, we can eliminate the effect of cluster sizes. Now, each cluster's score is already in the range [0,1], so we can take the average of each cluster's score as the result.

CI, CGI, and CRI each calculate a score for each cluster as a binary value of 0 or 1. The total score is the sum of these binary values in the range [0, $k$], providing an explainable measure. However, if desired, the score can also be normalized to the scale [0,1] by averaging (dividing the integer score by $k$) instead of summing. For example, relative-CI (referred to as *rel-CI*) has been used in [18]. In this case, the score is the percentage of the cluster errors, which is still explainable.

### 3.4 Implementation

ARI requires calculating the consistency of every pair of points, which may sound like a bit of a tedious process. Practical implementation uses a so-called *contingency table* of size $k^2$, where each cell [$a$, $b$] contains the count of points that belong to both cluster $a \in A$ and cluster $b \in B$. Each point contributes to one cell, and the contingency table can be created in O($n$) time. NMI and set-matching measures can also be computed from the contingency table. NMI calculates the probabilities and entropy whereas set-matching measures calculate the similarity of the clusters.

We have a C implementation of some of the measures in CBEVI[2], and the Python software of the ACC measure available on GitHub[3]. It includes two variants: a backward-compatible variant (ACC-pair) and a simpler variant (ACC-match) that avoids the Hungarian algorithm. The pseudo-code for ACC is below.

```
ACC-pair(La,Lb,mappingMethod):
kA = number of unique labels in La
kB = number of unique labels in Lb
N = length of La and Lb
contg = ContingencyMatrix(La,Lb,N,kA,kB,"pairing")
mapAtoB = hungarian(-contg)
return CountMatches(La,Lb,mapAtoB)
ACC-match(La,Lb):
acc1 = MatchOneWay(La,Lb)
acc2 = MatchOneWay(Lb,La)
return min(acc1,acc2)
MatchOneWay(La,Lb):
contg = ContingencyMatrix(La,Lb,N,kA,kB,"matching")
FOR i=1:kA
  mapAtoB[i] = ARGMAX{contg[i][j]: j=1,...,kB}
return CountMatches(La,Lb,mapAtoB)
CountMatches(La,Lb,N,mapAtoB):
correct=0
FOR i=1..N
  IF Lb[i] == mapAtoB[La[i]]
    correct+=1
return correct/N
```

[2] https://cs.uef.fi/ml/software/
[3] https://github.com/uef-machine-learning/ClusterAccuracy

```
ContingencyMatrix(La,Lb,N,kA,kB,mappingMethod)
IF mappingMethod == "pairing"
  contg = matrixOfZeros(MAX{kA,kB},MAX{kA,kB})
ELSE
  # Sparse matrix implemented using hash tables
  contg = matrixOfZeros(kA,kB)
FOR i=1..N
  # Number of points belonging to both clusters
  contg[La[i]][Lb[i]] += 1
```

## 4. Centroid index

NMI, ARI, ACC, and all the other measures in Section 3 are good for most practical use but lack in an important aspect: they are not explainable. A value in the range [0,1] is just a number where a higher value indicates a better clustering. But what does the value of 0.80 or 0.90 mean exactly? Should we aim at 0.99 or would 0.90 be sufficient?

Figure 9 shows the NMI and CI values for several clustering algorithms on seven selected datasets from [18]. Green indicates visually verified correct clustering results. NMI gives high values (0.93-1.00) for all the correct clustering. However, it sometimes provides high values also for incorrect clustering results. For example, the first column shows a Birch2 clustering result with 18 (out of 100) clusters incorrectly located (CI=18), yet a high NMI=0.96. In this case, the high values can be attributed to a known bias in NMI, especially for a large number of clusters ($k$=100) [36]. The problem is less severe in the $S_1$-$S_4$ datasets, but persists in two cases, with incorrect results of 0.93 and 0.95.

In other words, NMI results are difficult to interpret and not comparable across datasets. Centroid index (CI), on the other hand, provides an intuitive, and explainable value. The value directly indicates how many clusters are incorrectly solved. If all clusters are correctly solved, the centroid index provides a clear CI=0.

The main limitation of centroid index is that, as such, it can be used only for centroid-based clustering. However, a rather straightforward generalization to other clustering models has been proposed in [31]. The idea is to replace the nearest-centroid mapping by selecting the cluster with the most matches. Centroids are not needed. A slightly different variant called centroid ratio (CR) [30] counts the number of unstable centroids. Stability was calculated based on the distance between the matched centroids relative to the distance to the nearest centroids in the same clustering.

In general, centroid index and its variants (CI, CR, GCI) can be categorized as cluster-level measures. They measure cluster-level errors in the range [0, $k$] in contrast to point-level calculations of the other measures. This leads to a very intuitive, explainable measure.

Centroid index has a limitation: it does not even attempt to measure point-level differences. Figure 10 demonstrates the situation with two clustering results. The value CI=7 on the left ($Birch_2$) is clear: there are 7 places where the centroids (red) are in the wrong positions. At the same time, seven places miss a centroid (the red centroid is between two clusters). On the right ($S_3$), all centroids are at the correct places, matching perfectly with the ground truth. However, the centroids exhibit slight deviations, leading to point-level differences (yellow areas) that CI does not capture.

The point-level differences can therefore be needed in applications requiring measuring detailed point-level differences. A stability-based approach has been used to determine the number of clusters. However, it was shown to be highly sensitive to the parameter setup, to the extent that CI was found unsuitable as an evaluation measure [37].

**Normalized mutual information (NMI):**

| | | | | | | | | |
|---|---|---|---|---|---|---|---|---|
| *Birch*$_1$ | 0.95 | 0.97 | 0.99 | 0.96 | 0.98 | 1.00 | - | 1.00 |
| *Birch*$_2$ | 0.96 | 0.97 | 0.99 | 0.97 | 1.00 | 1.00 | - | 1.00 |
| *Birch*$_3$ | 0.90 | 0.94 | 0.94 | 0.93 | 0.93 | 0.96 | - | 1.00 |
| $S_1$ | **0.93** | 1.00 | 1.00 | 1.00 | 1.00 | 1.00 | 1.00 | 1.00 |
| $S_2$ | **0.90** | 0.99 | 0.99 | **0.95** | 0.99 | 0.93 | 0.99 | 0.99 |
| $S_3$ | **0.92** | 0.97 | 0.97 | 0.97 | 0.94 | 0.97 | 0.97 | 0.97 |
| $S_4$ | **0.88** | 0.94 | 0.94 | 0.95 | **0.85** | 0.94 | 0.94 | 0.94 |

**Centroid index (CI):**

| | | | | | | | | |
|---|---|---|---|---|---|---|---|---|
| *Birch*$_1$ | 7 | 3 | 1 | 4 | 0 | 0 | --- | 0 |
| *Birch*$_2$ | 18 | 11 | 4 | 12 | 0 | 0 | --- | 0 |
| *Birch*$_3$ | 23 | 11 | 7 | 10 | 7 | 2 | --- | 0 |
| $S_1$ | **2** | 0 | 0 | 0 | 0 | 0 | 0 | 0 |
| $S_2$ | **2** | 0 | 0 | **1** | 0 | 0 | 0 | 0 |
| $S_3$ | **1** | 0 | 0 | 0 | 0 | 0 | 0 | 0 |
| $S_4$ | **1** | 0 | 0 | 0 | **1** | 0 | 0 | 0 |

**Figure 9**: Sample results of NMI and CI. Each column (except the leftmost one) corresponds to an algorithm from [18], including k-means (the second column) and genetic algorithm (the rightmost column). Green indicates correct clusterings. Blue circles highlight inconsistent NMI results.

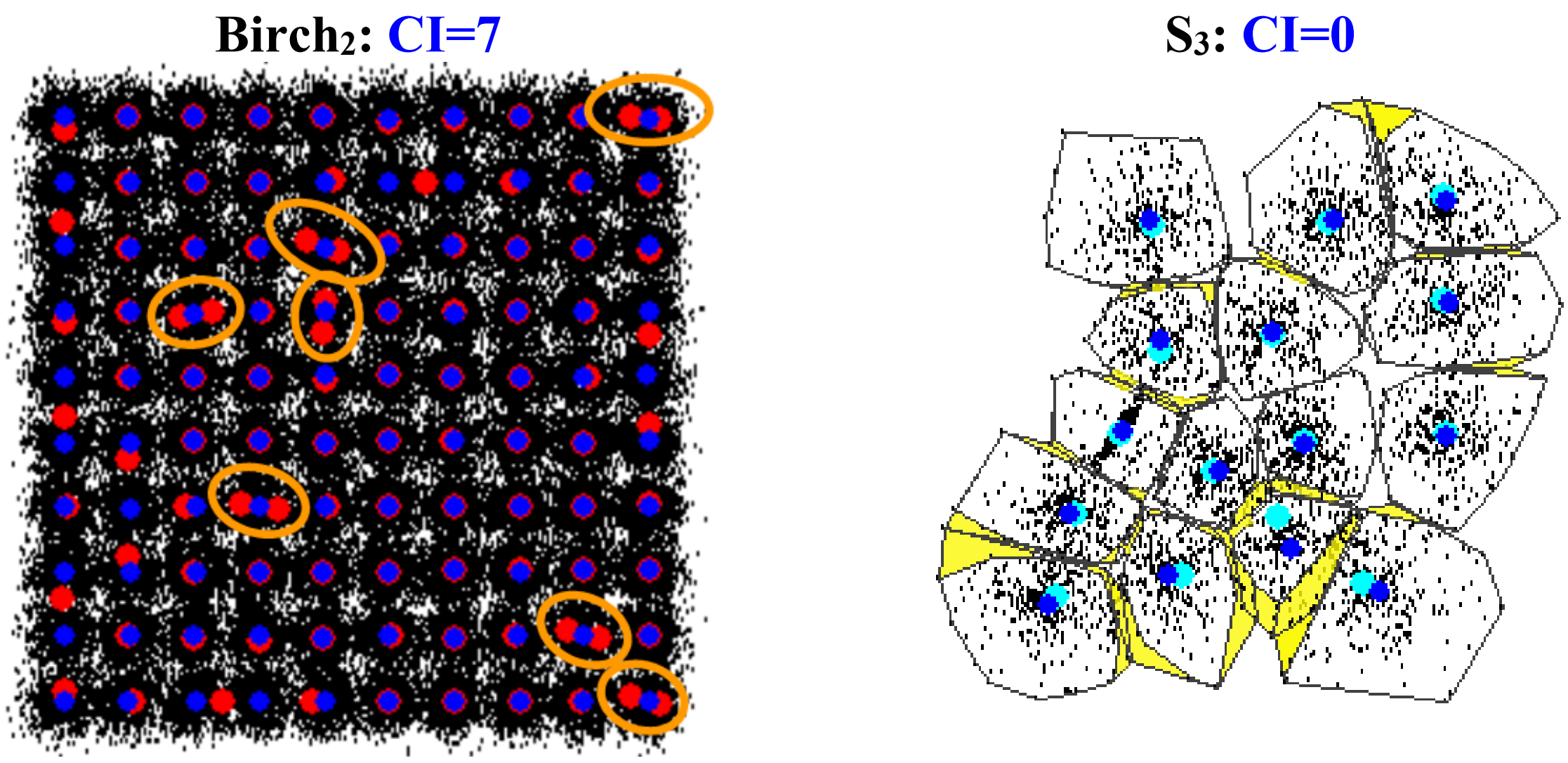

**Figure 10**: Two examples of centroid index. The CI=7 value (left) indicates that 7 centroids are mislocated. In other words, a real cluster has two centroids (orange circles) and one centroid tries to cover two real clusters in seven places. Value CI=0 (right) indicates that all centroids are roughly at their correct locations. However, CI does not measure point-level differences (yellow areas).

## 5. Conclusion and recommendations

We have reviewed clustering evaluation measures (external indexes) that use ground truth. They are needed when we want to compare the performance of clustering algorithms in a controlled environment.

We recommend using centroid index (CI) [5]. It is an intuitive cluster-level measure with an explainable result. If we need a fine-tuned point-level measure, our recommendation is clustering accuracy (ACC). It has been documented in detail, behavior analyzed, and an easy-to-use software is available [1].

For point-level measures, there are plenty of other choices. Namely, any other set-matching-based measure can be used. If a normalized score is needed, PSI [24] would be a better choice, as it does not favor larger clusters. However, if the argument is that all points should matter equally, then ACC or any other set-matching measure is good.

Among the pairwise consistency measures, ARI and NMI can also be used, but both have biases. In cases of unbalanced cluster sizes, ARI reflects the agreement among the largest clusters [23], whereas NMI is biased when there are many clusters [36].

The only measure we do not recommend is Rand index. It is biased regardless of the number of clusters or their sizes [22].